%% file: main.tex
\documentclass[sigconf]{acmart}
\AtBeginDocument{%
  }

\usepackage{amssymb}
\usepackage{graphicx}

\setcopyright{acmlicensed}
\copyrightyear{2026}
\acmYear{2026}
\setcopyright{cc}
\setcctype{by}
\acmConference[MM '26]{Proceedings of the 35th ACM International Conference on Multimedia}{November 10--14, 2026}{Rio de Janeiro, Brazil}
\acmBooktitle{Proceedings of the 35th ACM International Conference on Multimedia (MM '26), November 10--14, 2026, Rio de Janeiro, Brazil}
\acmDOI{10.1145/3767308.3836272}
\acmISBN{979-8-4007-2213-4/2026/11}
\usepackage{multirow}
\usepackage{booktabs}
\usepackage{tcolorbox}
\newtcolorbox{promptbox}{
    colback=gray!10,
    colframe=gray!40,
    sharp corners
}
\acmSubmissionID{123-A56-BU3}

\begin{document}

\title{SEE: Structure-aware Exploring \& Exploiting for Long-horizon GUI Agent Trajectory Synthesis}




\author{Zhuohang Fan}
\email{25b903062@stu.hit.edu.cn}
\orcid{0009-0005-6159-0117}
\affiliation{%
  \institution{Harbin Institute of Technology}
  \city{Weihai}
  \country{China}
}

\author{Beichen Zhang}
\email{beiczhang@hit.edu.cn}
\authornote{Corresponding author.}
\orcid{0000-0001-5030-0632}
\affiliation{%
  \institution{Harbin Institute of Technology}
  \city{Weihai}
  \country{China}
}

\author{Yuanfa Li}
\email{ifli2023@gmail.com}
\orcid{0009-0003-0754-9039}
\affiliation{%
  \institution{Researcher}
  \city{Beijing} 
  \country{China}
}

\author{Changqiao Wu}
\email{changqiao.wu@gmail.com}
\orcid{0009-0003-3533-422X }
\affiliation{%
  \institution{Researcher}
  \city{Beijing} 
  \country{China}
}

\author{Wei Liu}
\email{buptliuwei@gmail.com}
\orcid{0009-0009-4327-1920}
\affiliation{%
  \institution{Researcher}
  \city{Beijing} 
  \country{China}
}

\author{Jian Luan}
\email{jian.luan@gmail.com}
\orcid{0000-0002-2383-226X}
\affiliation{%
  \institution{Researcher}
  \city{Beijing} 
  \country{China}
}

\author{Weigang Zhang}
\email{wgzhang@hit.edu.cn}
\authornote{Corresponding author.}
\orcid{0000-0003-0042-7074}
\affiliation{%
  \institution{Harbin Institute of Technology}
  \city{Weihai}
  \country{China},
  }
\affiliation{
  \institution{Qingdao Research Institute}
  \city{Qingdao}
  \country{China},
}






\begin{abstract}
Graphical User Interface (GUI) agents powered by vision-language models hold promise for automating real-world mobile tasks. However, progress is limited by the lack of high-coverage, long-horizon interaction trajectories collected from element-rich and rapidly evolving apps. Existing pipelines often rely on costly human demonstrations or on-policy framework, which tends to over-sample common flows while missing rare transitions and complex multi-step procedures. To address this problem, we propose SEE, a two-stage data synthesis framework consisting of \textit{(i)} an efficient exploration stage that builds an explicit UI transition graph over screens and elements, and \textit{(ii)} a graph-based synthesis stage that composes diverse multi-step trajectories via planning and controlled sampling. This design yields reproducible and explainable data generation, while explicitly preventing spurious cycles and enabling long-horizon composition. Across multiple real-world apps, SEE produces trajectories with an average length of \textbf{14.8} steps while avoiding spurious loops, and agents fine-tuned on SEE achieve improved task success and generalization to unseen screens. Project page \href{https://github.com/Zhh-Fan/SEE-Structure-aware-Exploring-Exploiting-for-Long-horizon-GUI-Agent-Trajectory-Synthesis}{is here}.
\end{abstract}



\maketitle

\vspace{-5mm}
\section{Introduction}
\begin{figure*}[!t]
    \centering
    \includegraphics[width=0.9\linewidth, trim=0 0 0 0, clip]{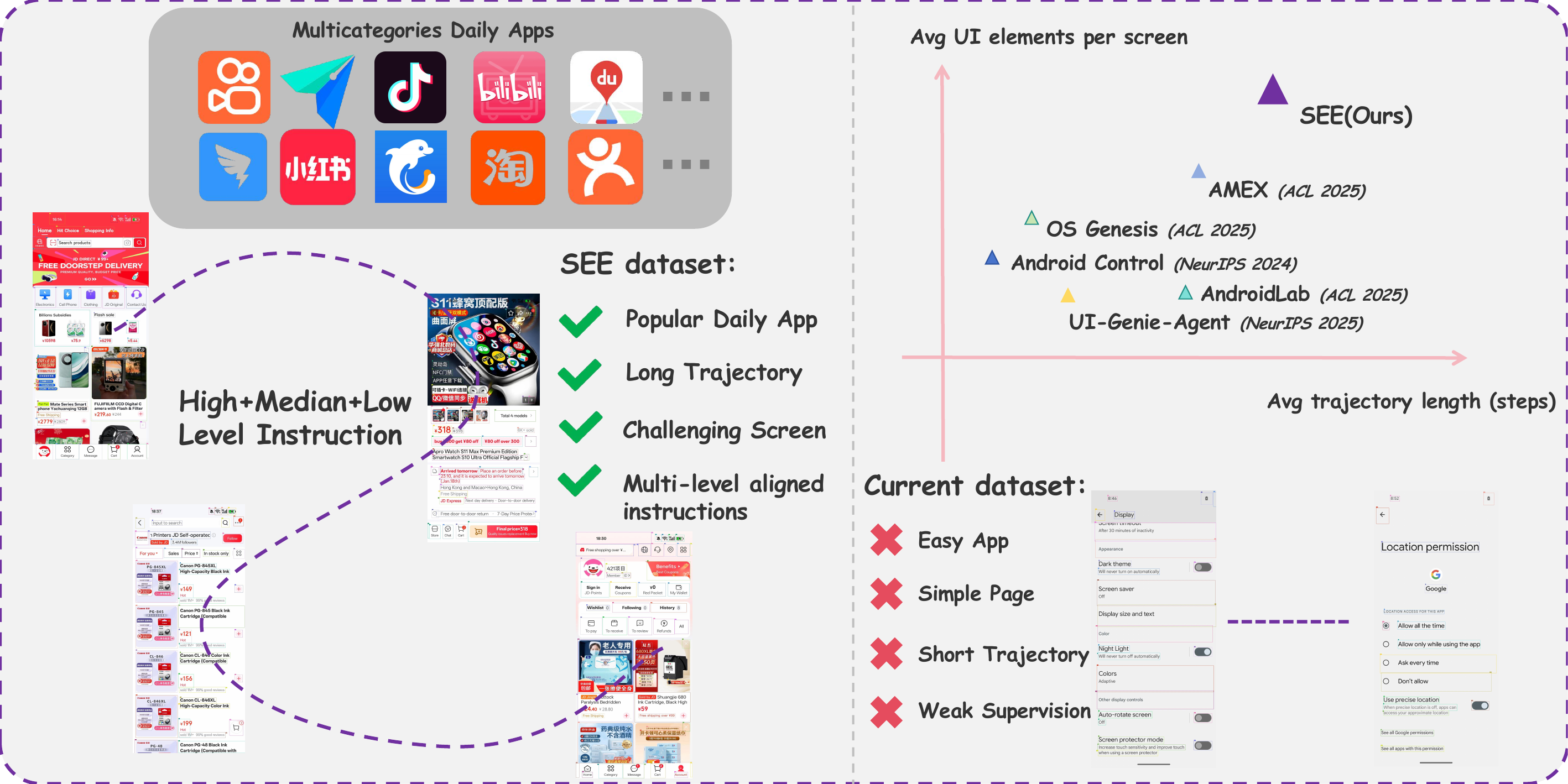}
    \caption{SEE targets the long-horizon, high-complexity regime of real-world mobile apps. \textbf{Left:} our dataset is built from popular, fast-evolving mobile applications and provides multi-level, step-aligned supervision.
\textbf{Right:} comparison across public datasets in the plane of average trajectory length and page complexity. SEE occupies the high-length, high-complexity region. 
}
\Description{Comparison of SEE dataset.}
    \label{fig:complexity_steps}
    \vspace{-2mm}
\end{figure*}

Vision--Language Models (VLMs) are rapidly advancing autonomous agents in complex digital environments~\cite{zhang2025largelanguagemodelbrainedgui,wang2025guiagentsfoundationmodels}. One particularly important application is GUI agents, where LLM-based controllers interact with graphical user interfaces to complete real-world tasks. Although recent systems have shown promising results on controlled benchmarks and simplified interfaces~\cite{hong2024cogagent,huagents}, they remain fragile in realistic mobile settings. Modern mobile apps are visually crowded, heavily dynamic, and frequently updated. They contain dense and repetitive UI elements, transient overlays such as permissions and pop-ups, and prerequisite-dependent flows. 
In addition, real users often require long-horizon procedures that involve interruptions, recovery from detours, and navigation across diverse interface states.
Consequently, even as many works improve planning and policy learning for long-horizon interaction, progress is still fundamentally constrained by the lack of scalable trajectory data that matches the complexity of real-world applications.

Existing GUI agent datasets and synthesis pipelines largely operate in a simplified regime.
Most are collected from relatively static apps or simple screens, and their trajectories are dominated by short, “happy-path” interactions with limited coverage of long-tail states and rare transitions. As illustrated in Figure~\ref{fig:complexity_steps}, public datasets tend to concentrate in either the short-horizon or the low-complexity region, leaving the long-horizon, element-rich quadrant largely uncovered. This mismatch is increasingly problematic because modern GUI agents must ground actions among many visually similar elements, handle transient interruptions, and execute multi-step procedures that depend on prior navigation history. Current data therefore provides insufficient support for training robust agents in realistic mobile environments.
 

To close this gap, we draw inspiration from Human Computer Interaction (HCI) and cognitive studies, which describe GUI interaction as a goal-driven action cycle: users form goals, specify intentions and actions, execute, perceive outcomes, and evaluate progress---often iterating via correction and backtracking \cite{zhang2024inductive,norman2013design,10.1145/235833.236050,10.1145/358886.358895,jaimes2007multimodal}. This suggests a natural decomposition of data generation into two complementary modes: \textit{(i)} perceptual exploration to probe affordances and map what is reachable, and \textit{(ii)} structure-aware exploitation to compose purposeful multi-step procedures toward user goals.

Building on this insight, we propose \textbf{SEE}, a \textbf{S}tructure-aware \textbf{E}xploring and \textbf{E}xploiting framework for GUI-agent data generation. In the exploration stage, the agent combines visual parsing, semantic priors, and short-term action history to efficiently discover meaningful interface states while constructing an explicit screen-element transition graph.
In the synthesis stage, SEE generates long-horizon trajectories by planning over this graph rather than repeatedly re-interpreting raw screenshots. Specifically, it first selects target screens and organizes them into an ordered subgoal sequence, and then obtains the corresponding executable action trace through graph search.
This design yields controllable long-horizon generation since both the number of subgoals and their graph distances can be used to adjust task length and difficulty. Moreover, by reusing verified node and edge semantics, SEE supports scalable and more cost-stable synthesis of executable trajectories with aligned multi-level supervision.



In summary, our main contributions are as follows: \textbf{(1)} We propose a structure-aware exploration framework that builds a screen-element transition graph for complex mobile apps. \textbf{(2)} We introduce a graph-based synthesis pipeline that generates executable long-horizon trajectories with multi-level language supervision. \textbf{(3)} We construct SEE, a GUI dataset with longer trajectories and denser screens than prior public datasets, and show that it improves downstream agent performance and cross-benchmark transfer.

\begin{figure*}[t]
    \centering
    \includegraphics[width=0.9\linewidth, trim=0 12 0 18,clip]{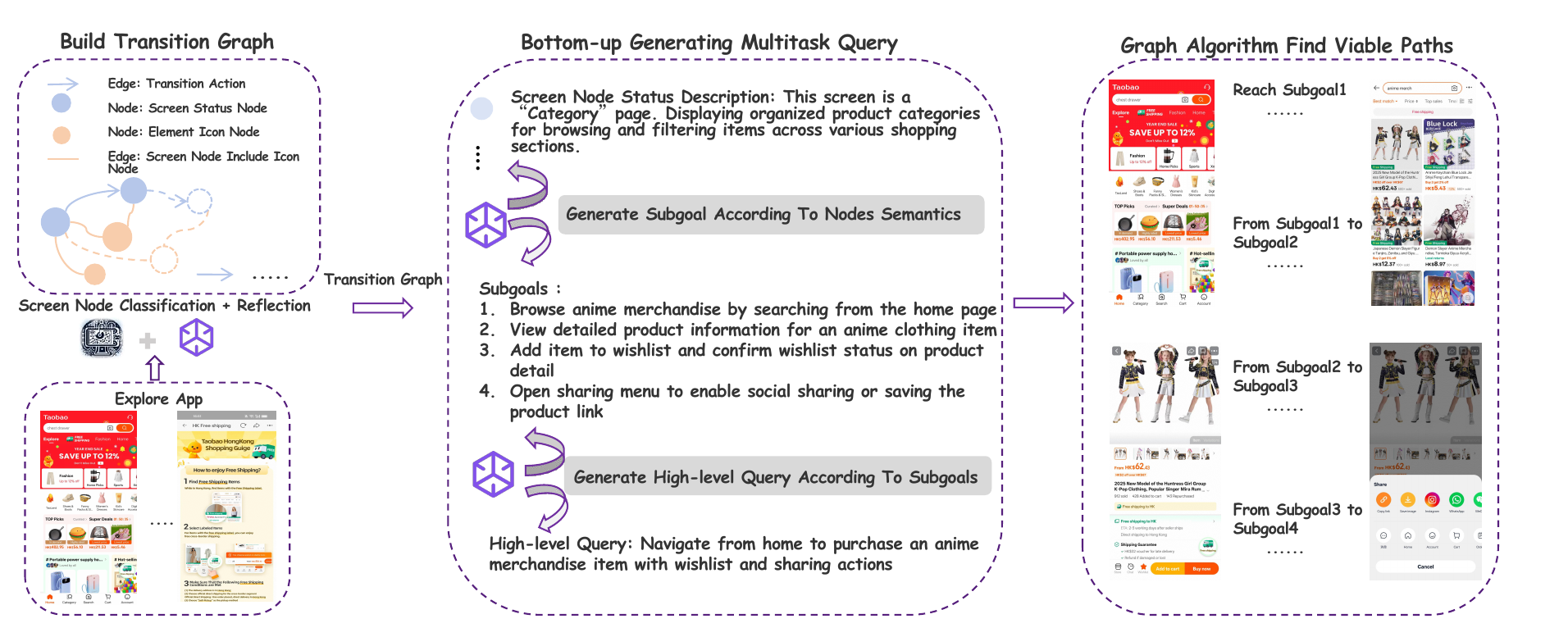}
    \vspace{-2mm}
    \caption{Overview of SEE. In the exploration stage, the agent interacts with the app to build a screen–element transition graph. In the synthesis stage, SEE generates high-level tasks and executable long-horizon trajectories by planning over the discovered graph.}
    \label{fig:Framework}
    \Description{Overview of SEE.}
    \vspace{-2mm}
\end{figure*}
\section{Related Work}

\paragraph{GUI agents.}
GUI agents have evolved from rule-based automation to data-driven end-to-end systems. Early RPA-style tools and benchmarks such as DART and WoB relied on handcrafted rules or scripted UI actions, which limited robustness to interface variation and reduced generalization to unseen layouts \citep{10.1145/3511667,repec,1235451,pmlr-v70-shi17a}. With the emergence of foundation models, modular agent systems that combine perception, memory, grounding, and tool use have enabled more adaptive multi-step interaction. However, such modular pipelines are often sensitive to prompting and brittle in their coordination across components~\citep{zhang2020state,openai2024gpt4technicalreport,tan2024towards,yan2023gpt4vwonderlandlargemultimodal,yang2023autogptonlinedecisionmaking,xia2024agentlessdemystifyingllmbasedsoftware}. More recent native GUI agents move toward unified vision-centric policies for web, mobile, and desktop interaction, supported by specialized datasets and training strategies \citep{anthropic2024claude,xu2025aguvis,wuatlas,qin2025uitarspioneeringautomatedgui}. These advances have substantially improved the capability of GUI agents, but their performance still depends heavily on the quality and coverage of training trajectories.

\paragraph{Benchmarks and environments.}
Benchmarks have progressed from controlled web environments (MiniWob++, WoB) to more realistic and long-horizon settings, including WebShop, Mind2Web, WebArena, VisualWebArena, and WebLINX~\citep{liu2018reinforcement,pmlr-v70-shi17a,NEURIPS2022_82ad13ec,NEURIPS2023_5950bf29,koh2024visualwebarena,lu2024weblinx}. Beyond web environments, OSWorld and WindowsAgentArena support full desktop workflows, while benchmarks like AndroidWorld, GUI-Odyssey, AndroidControl expand to mobile within- and cross-app interaction \citep{NEURIPS2024_5d413e48,pmlr-v267-bonatti25a,rawlesandroidworld,lu2025guiodyssey}. 

\paragraph{Data collection and synthesis.}
Large-scale GUI training data has traditionally relied on human demonstrations~\cite{zhang2024downstream}, as in AndroidControl, WebShop, Mind2Web, and GUI-Odyssey \citep{NEURIPS2022_82ad13ec,NEURIPS2023_5950bf29,NEURIPS2024_a79f3ef3,lu2025guiodyssey}. Although such data is high-fidelity, it is expensive and difficult to scale. Recent work explores automated synthesis via exploration or search, including methods such as OS-Genesis and WebSynthesis, as well as alternative supervision sources such as transformed external resources or video-grounded interaction data~\citep{sun2025genesis,gao2025websynthesisworldmodelguidedmctsefficient,zhang2026tongui,chengui}. Recent research like GUI-ReWalk and UI-Genie further study scalable synthetic trajectory construction for GUI agents \citep{lin2025guirewalkmassivedatageneration,xiaoui}. 
Overall, existing GUI data pipelines can be broadly grouped into three categories: demonstration-based collection, online exploration or search-based synthesis, and graph- or structure-informed generation. SEE differs from previous work in three ways. First, unlike demonstration-heavy approaches, SEE constructs the transition graph without manual trajectory annotation. Second, unlike online synthesis pipelines that repeatedly generate trajectories through step-by-step reasoning or search, SEE explicitly decouples structure acquisition from trajectory construction, and synthesizes executable trajectories by planning over verified graph semantics. Third, SEE introduces reflection-based graph refinement before large-scale synthesis, which improves transition reliability and helps avoid noisy or spurious edges. 

\section{Structure-aware Exploring \& Exploiting (SEE)}

In this section, we present SEE, a two-stage framework for generating GUI-agent training data. SEE first acquires an explicit structural representation of an app through structure-aware exploration, and then synthesizes executable long-horizon trajectories by planning over the discovered graph. As illustrated in Figure~\ref{fig:Framework}, the framework consists of three main components: structure-aware exploration, graph refinement, and graph-based trajectory synthesis.

\subsection{Preliminaries}
\label{sec:prelim}

We model interacting within applications as a partially observable sequential decision process.
Let $u_t \in \mathcal{U}$ denote the abstract UI state at step $t$, and let
$o_t \in \mathcal{O}$ denote the corresponding observation. At step $t$, the agent executes an action
$a_t \in \mathcal{A}$, where $a_t = (\mathrm{op}_t, \mathrm{arg}_t)$ consists of an
operation type and its execution arguments. The environment evolves according to
\begin{equation}
u_{t+1} \sim P(\cdot \mid u_t, a_t), \qquad
\end{equation}
where $P$ is the state transition model.
An exploration trajectory is formulated as
\begin{equation}
\mathcal \tau = \bigl((o_0,a_0),(o_1,a_1),\ldots,(o_{L-1},a_{L-1}),o_L\bigr),
\end{equation}
with horizon $L$. During exploration, the Back action is triggered by a simple rule when the agent reaches a deep screen that yields no useful new transitions. This mechanism allows the agent to return to previously visited states and continue exploring other branches. During synthesis, by contrast, Back is treated as a normal planned action on the transition graph, since it may be necessary for reaching later subgoals.



\paragraph{UI state abstraction and transition graph.}
To support structured trajectory generation, we introduce an abstraction function $\phi:\mathcal{O}\rightarrow\mathcal{U}$ that
maps observations to discrete UI states $u_t=\phi(o_t)$.

We then represent an app by a graph $G=(V,E)$, where the node set $V$ contains two types of nodes: screen nodes $u\in\mathcal{U}$, which represent UI states, and element nodes $e\in\mathcal{M}$, which represent UI elements with semantic descriptions and parsed attributes. The edge set $E$ contains two types of edges: (i) transition edges $(u,a,u')$, which encode executable action-induced transitions between screens, and (ii) containment edges $(u,e)$, which indicate which elements appear on a given screen. This bipartite screen--element graph captures both navigational structure and screen-level semantic affordances.

\begin{figure*}[t]
    \centering
    \includegraphics[width=0.95\linewidth, clip, trim=0 15 0 10]{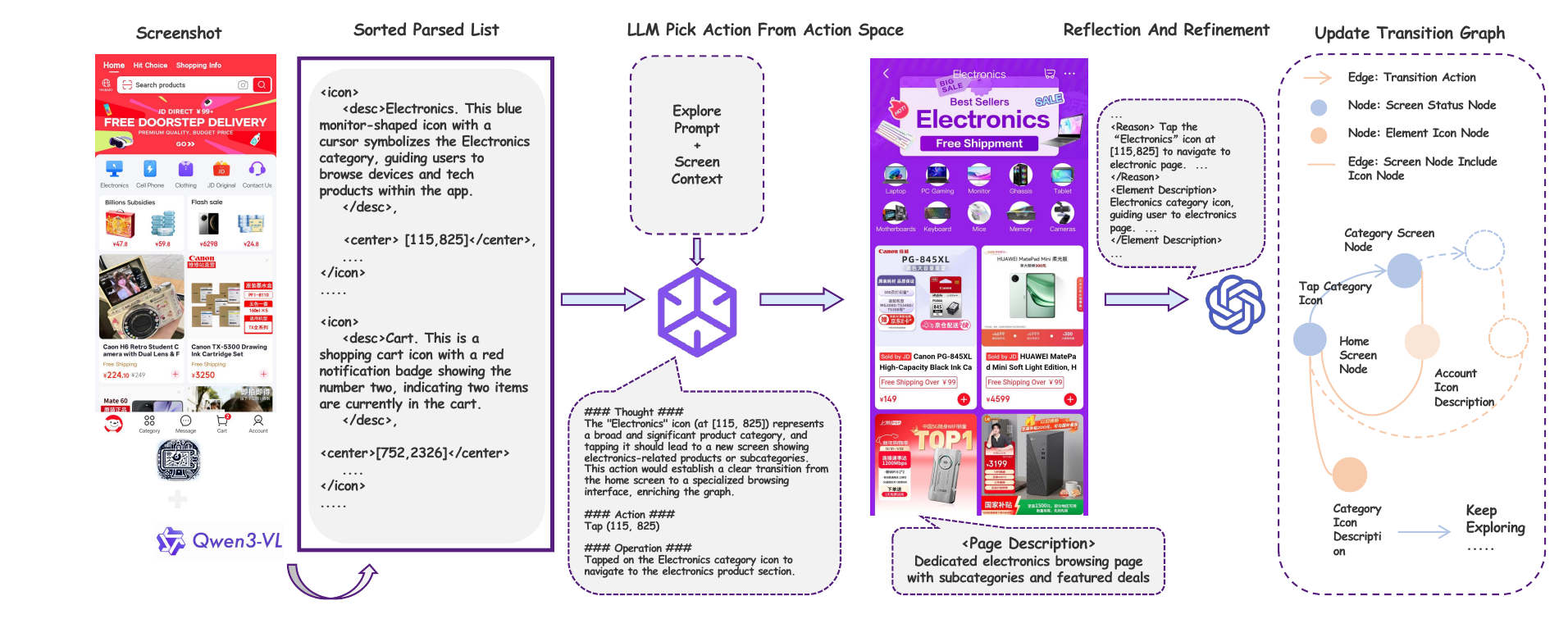}
    \vspace{-6mm}
    \caption{Vision-and-semantics guided exploration for transition graph construction. Candidate UI elements are parsed and semantically described, an action is selected to probe a meaningful transition, and the resulting observation is used to update the transition graph with verified nodes, edges, and refined semantics.}
    \label{fig:Exploration}
    \Description{Vision-and-semantics guided exploration for transition graph construction.}
    \vspace{-2mm}
\end{figure*}

\subsection{Structure-aware Exploration}
\label{sec: exploration}


The goal of the exploration stage is to discover diverse and meaningful UI states while constructing the transition graph $G=(V,E)$. Instead of relying on naive trial-and-error interaction, SEE guides exploration with structural priors, semantic cues, and short-horizon interaction history. This design encourages the agent to visit functionally important states and collect transitions that are useful for later graph-based synthesis.

Given the current observation \(o_t\), we first apply a UI parser~\cite{wan2024omniparser} to extract a set of candidate elements $\mathcal{M}_t=\{e_{t,i}\}_{i=1}^{N_t}$, together with their basic attributes, such as location and element type. Because many mobile interfaces contain complex icons or visually ambiguous controls, we further use a vision-language model to generate short semantic descriptions for these elements. These descriptions allow the agent to reason about element functionality rather than relying solely on raw appearance.


\subsubsection{Exploration action selection.} We then rank the candidate elements according to their relevance to the current app context. The ranking score combines three factors: (i) semantic alignment with app-category and navigation priors, (ii) a lightweight layout prior that favors structurally important regions such as the top or bottom bars, and (iii) history-based penalties that discourage repeatedly selecting uninformative or redundant elements:
\begin{equation}
s(e_{t,i}) = s_{\text{sem}}(e_{t,i}) + s_{\text{lay}}(e_{t,i}) - p_{\text{hist}}(e_{t,i}) 
\end{equation}
Here, \(s_{\text{sem}}\) measures semantic relevance to app's category, \(s_{\text{lay}}\) captures layout preference, and \(p_{\text{hist}}\) penalize repeated or near-duplicate choices. We retain the top-\(K\) candidates and let the language model select an action from this reduced set. To improve branch coverage, we also introduce a small random exploration probability that occasionally allows the agent to move beyond the highest-ranked candidates.

\subsubsection{Structure-aware screen identification.}
To update the graph $G$ consistently, we map each screen observation $o_t$  to a discrete screen state
$u_t=\phi(o_t)$ through visual matching, analogous to human perceptual processing.
For each observation, we extract a set of $N$ elements from predefined regions of interest
$\mathcal{R}$, such as the top and bottom bars, which are usually more stable across interactions. Each prototype is represented as
$\mathcal{P}(o)=\{p_i\}_{i=1}^{N}$ with $p_i=(f_i,b_i)$,
where $f_i\in\mathbb{R}^{d}$ is a visual embedding obtained by MobileNet~\cite{howard2019searching} and $b_i$ is its bounding box.

Given two observations $o$ and $o'$, we compute a pairwise cost matrix as follows,
\begin{equation}
C_{ij}(o,o')=\alpha\, d_{\cos}(f_i,f'_j)+\beta\, d_{\text{pos}}(b_i,b'_j),
\end{equation}
where $d_{\cos}$ measures the visual difference between two elements in the feature space, and $d_{\text{pos}}$ measures the difference between their spatial positions. A lower cost means that the two elements are more similar both in appearance and in layout. The coefficients $\alpha$ and $\beta$ control the relative importance of visual similarity and positional consistency.
We then use Hungarian matching to find the optimal one-to-one alignment between the representative elements in the two observations. If a sufficient number of aligned pairs have low matching cost, we regard $o$ and $o'$ as the same screen state. 

\subsection{Graph Update with Reflection and Refinement}

Because the transition graph serves as the structural foundation of the synthesis stage, we perform reflection and refinement after each executed action to improve the quality of both nodes and edges in the constructed graph $G$.

After an action is executed, SEE runs a reflection step to verify whether the observed outcome is consistent with the semantics of the selected element and the intended action. In particular, the reflection record indicates whether the transition is successful, whether it leads to a meaningful new state, and whether the resulting screen is semantically consistent with the function suggested by the acted-on element. Only the verified transition edge will be added to the transition graph $G$.

In parallel, we refine the semantic description of the interacted element based on the observed outcome. This refinement reduces ambiguity in the early-stage exploration and improves the semantic quality of later synthesized trajectories. After verification, we update the graph by adding the verified transition edge $(u_t, a_t, u_{t+1})$ together with the corresponding containment edges $(u_t \rightarrow e)$ for the relevant refined interacted elements $e \in \mathcal{M}$.

Importantly, we do not discard system interruptions such as permission dialogs, service agreements, or notification overlays. Instead, we record these cases and their recovery paths as normal transitions. This allows the synthesized trajectories to include interruption-handling behaviors that commonly arise in real-world mobile interaction. Please see supplementary material and Table~\ref{tab:graph_quality_multi_scale} for more details.

\begin{table*}[ht]
  \begin{center}
    \begin{small}
      \begin{sc}
      
        \begin{tabular}{lccccr}
          \toprule
          \textbf{Datasets} &\textbf{Size}  & \textbf{Manual} &\textbf{Average} &\textbf{Trajectories} & \textbf{Task} \\
        & & \textbf{Annotation} & \textbf{Steps} & & \textbf{Instruct} \\
          \midrule
          Android Control    & 88K & $\surd$  & 5.5 & 15283 & High $\&$ Low \\
          AMEX & 37K & $\surd$  & 12.8 & 2946 & High \\
          Androidlab    & 6K & $\surd$ & 8.6 & 726 & High\\
          UI-GENIE-Agent-16K      & 16K & $\times$ & 7.1 & 2208 & High $\&$ Low   \\
          OS-Genesis & 16K & $\times$ & 6.4 & 2451 & High $\&$ Low\\ 
          \midrule
          \textbf{SEE-Train}   & 47K & $\times$ & 14.8 &  3237 & High $\&$ Median $\&$ Low\\
          \textbf{SEE-Test}   & 5K & $\times$ & 12.9 &  419 & High $\&$ Median $\&$ Low\\
          \bottomrule
        \end{tabular}
      \end{sc}
    \end{small}
  \end{center}
  \caption{Comparison of SEE data and other GUI datasets.}
  
  \vskip -0.3in
\label{tab:data_stats}
\end{table*}

\subsection{Graph Based Data Synthesis}
\label{sec:synthesis}

Given the constructed transition graph $G=(V,E)$, we synthesize executable trajectories by composing graph paths on $G$ under controllable constraints. Each synthesized episode contains a high-level instruction, an ordered subgoal list, and aligned step-level action descriptions.

\subsubsection{Generating subgoals and high-level instructions.}
Each screen node $u\in\mathcal{U}$ is associated with a concise semantic state description produced during exploration (Sec.~\ref{sec: exploration}). To construct a multi-step task, we first sample a set of candidate screens from $V$ using a controllable policy that can emphasize long-tail nodes and diverse functional regions. Conditioned on the sampled screen semantics, an LLM generates an ordered subgoal list $\mathcal{L}$ and summarizes it into a high-level instruction that defines the overall episode task intent.

\subsubsection{Graph-based trajectory composition.}
Given the generated subgoals $\{l_j\}$ and their associated target screen nodes $\{u^{(j)}\}$, SEE synthesizes an executable trajectory by connecting these targets on the constructed graph. Concretely, for each consecutive pair $(u^{(j)},u^{(j+1)})$, we run a graph search algorithm, such as BFS, and concatenate the resulting transition paths into a complete action sequence. Because the synthesis operates on an explicit transition structure, SEE can directly control (i) trajectory length via the number of subgoals and the selected path lengths, and
(ii) coverage by adjusting how screen nodes are sampled during the task construction phase. We can also apply lightweight constraints during path selection to avoid spurious cycles and redundant oscillations.

\begin{table}[]
\centering
\small
\setlength{\tabcolsep}{10pt}
\begin{tabular}{lccc}
\toprule
\textbf{Subgoals range} & \textbf{Ratio} & \textbf{Avg steps} \\
\midrule
$1$--$3$  & 12.41\% & 6.98 \\
$4$--$6$  & 64.04\% & 14.51 \\
$\ge 7$   & 23.09\% & 20.16 \\
\bottomrule
\end{tabular}
\vspace{1mm}
\caption{\textbf{Trajectory length and subgoal count.}}
\label{tab:subgoal_bins_steps}
\vskip -0.3in
\end{table}

\begin{table}[]
\centering
\small
\setlength{\tabcolsep}{10pt}
\begin{tabular}{lc}
\toprule
\textbf{Dataset} & \textbf{Avg Elements / Screen} $\uparrow$ \\
\midrule
AndroidLAB  &  13.95\\
UI-Genie-Agent-16K  & 13.59 \\
AndroidControl  & 13.83  \\
AMEX & 16.60 \\
GUI-Xplore &  17.26\\
OS-Genesis & 14.47 \\
\midrule
\textbf{Ours} & \textbf{24.63} \\
\bottomrule
\end{tabular}

\caption{Page complexity comparison across datasets.}
\label{tab:ui_complexity_simple}
\vskip -0.3in
\end{table}

\subsubsection{Low-level instruction synthesis from node/edge semantics.}

Beyond the executable action trace, we generate low-level step descriptions that specify what to do at each step $t$ and what outcome is expected. A key advantage of SEE is that these descriptions can be synthesized economically and precisely by using the semantic description stored in the transition graph, rather than re-interpreting raw screenshots at every step. Concretely, during exploration each transition edge $(u,a,u')$ is associated with a refined textual explanation that summarizes the observed transition (e.g., Fig.~\ref{fig:Exploration}). In addition, each screen node maintains a concise semantic state summary, and each element node stores a short functional description.
During synthesis, for a selected graph path $(u_0 \xrightarrow{a_0} u_1 \xrightarrow{a_1} \cdots \xrightarrow{a_{L-1}} u_L)$, we prompt a language model to render the low-level instruction for step $t$ by conditioning on: (i) the current screen summary of $u_t$, (ii) the interacted element semantics (if applicable), and (iii) the edge-level refinement attached to $(u_t,a_t,u_{t+1})$, which provides a verified description of the expected transition outcome. 


Unlike many other LLM-driven data generation pipelines, which require step-by-step model decisions for every synthesized trajectory, SEE decouples structure acquisition from trajectory construction: once the graph is built, the action sequence is largely determined by graph paths, and low-level instructions are generated by reusing node and edge annotations rather than repeatedly parsing screenshots or re-planning. As a result, the marginal token cost of adding new trajectories is dominated by lightweight subgoal prompting and short text rendering, making large-scale synthesis more cost-stable even for long-horizon workflows.
Overall, each episode in SEE consists of an executable GUI trajectory paired with multi-level language supervision. Concretely, the dataset includes a high-level task instruction $I$, an ordered subgoal sequence $\mathcal{L}$, and each step is accompanied by a fine-grained action description.

\begin{table*}[ht]
\centering
\small
\setlength{\tabcolsep}{5pt}
\begin{tabular}{lccc ccccc}
\toprule
\multirow{2}{*}{\textbf{Agent}} &
\multirow{2}{*}{\textbf{SR} $\uparrow$} &
\multirow{2}{*}{\textbf{Grounding} $\uparrow$} &
\multirow{2}{*}{\textbf{Type} $\uparrow$} &
\multirow{2}{*}{\textbf{Model Size}} &
\multicolumn{4}{c}{\textbf{Step Accuracy Breakdown} $\uparrow$} \\
\cmidrule(lr){6-9}
& & & & & \textbf{Tap} & \textbf{Swipe} & \textbf{Back} & \textbf{Type(text)} \\
\midrule
Qwen2.5-VL & 39.92\% & 42.76\% & 68.83\% & 7B & 37.21\% & 73.07\% & 34.05\% & 63.47\% \\
UI-Genie(Base Qwen2.5-vl)  & 45.77\% & 44.01\% & 77.59\% & 7B & 36.60\%  & 69.51\% & 69.51\% & 63.47\%  \\
GUI-Owl-7B(Base Qwen2.5-vl)  & 45.94\% & 47.04\% & 76.70\% & 7B & 39.21\% & 66.67\% & 59.76\% & 68.97\% \\
OS-Genesis-7B(Base  Qwen2-vl) & 45.17\% & 40.52\% & 67.58\% & 7B & 39.29\% & 63.33\% & 58.76\% & 72.41\% \\
OS-Atlas-7B(Base Qwen2-vl) & 53.32\% & 49.38\% & 69.89\% & 7B & 50.44\% & 64.33\% & 63.17\% & 73.33\% \\
Qwen3-VL & 62.61\% & 60.79\% & 93.04\% & 4B & 54.76\% & 81.10\% & 85.43\% & 73.45\%  \\
\midrule
\textbf{Qwen3-VL (Train on SEE-Train)} & \textbf{77.29\%} & \textbf{69.91\%} & \textbf{97.87\%} & 4B & \textbf{67.16\%} & \textbf{97.20\%} & \textbf{97.76\%} & \textbf{96.67\%} \\
\bottomrule
\end{tabular}
\vspace{2mm}
\caption{Benchmarking GUI agents on SEE-Test.
We report Success Rate (SR), together with per-step breakdown (Tap/Swipe/Back/Type), UI Element Interaction Accuracy (Grounding), and overall Action Type Accuracy (Type).}
\label{tab:main_agents}
\vspace{-5mm}
\end{table*}

\section{SEE Data Statistics} 

SEE is synthesized without manual trajectory annotation, yet provides multi-level supervision for each episode:
a high-level instruction, an ordered list of subgoals (median-level), and step-aligned fine-grained action descriptions (low-level). Importantly, SEE-Test consists of trajectories collected from apps that are disjoint from SEE-Train, rather than reused or sampled from the training split.

\noindent \textbf{Overall Statistics.} Table~\ref{tab:data_stats} compares SEE with other representative public GUI datasets.
Compared with demonstration-heavy datasets such as Android Control, AMEX and AndroidLab~\cite{Chai_2025, xu2025androidlab, NEURIPS2024_a79f3ef3}, SEE removes the manual annotation while still maintaining executable trajectories.
Compared with recent synthetic pipelines such as UI-GENIE and OS-Genesis ~\cite{xiaoui, sun2025genesis}, SEE emphasizes long-horizon procedures.
Although the number of trajectories is comparable to several recent synthetic datasets, SEE reaches an average trajectory length of 14.8 steps, which is substantially longer than datasets dominated by short, single-screen interactions. In addition, SEE provides aligned supervision at the high-, median-, and step-level, which is particularly important for realistic mobile tasks that require multiple intermediate goals and repeated screen transitions.

\label{sec:stats}
\begin{figure}
    \centering
    \vskip -0.12in
    \includegraphics[width=0.9\linewidth]{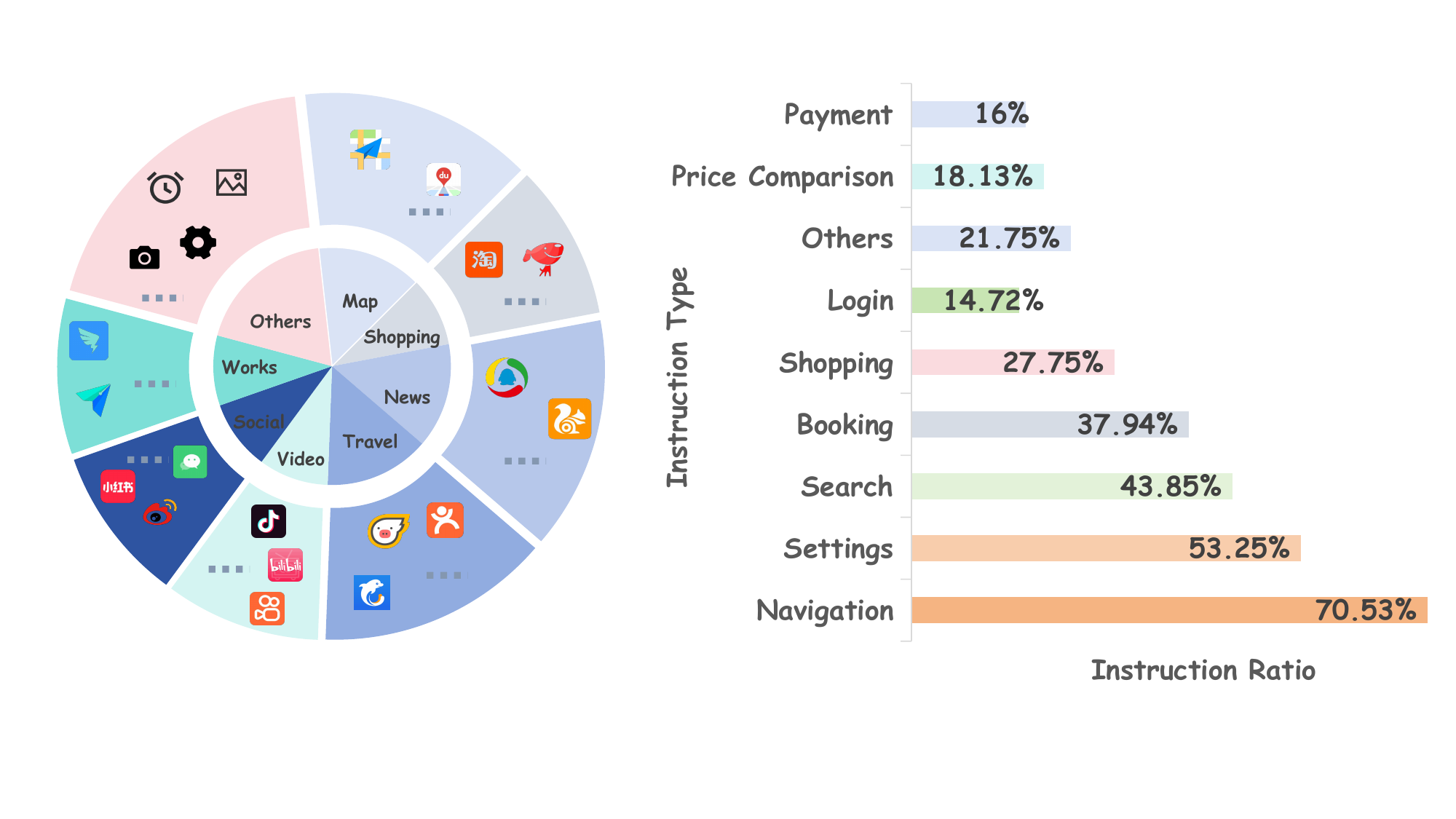}
    \vspace{-10mm}
    \caption{Dataset coverage and instruction composition.}
    \Description{Dataset coverage and instruction composition in SEE.}
    \label{fig:coverage_and_types}
\vspace{-6mm}
\end{figure}

\noindent \textbf{Trajectory length.} To better characterize the compositional structure of SEE, we analyze the number of subgoals per episode and its relationship to trajectory length. As shown in Table~\ref{tab:subgoal_bins_steps}, SEE is dominated by multi-stage instructions: episodes with $4$--$6$ subgoals accounts for the majority of episodes with an average of 14.51 steps, while a substantial portion of episodes containing at least $\ge 7$ subgoals require even longer execution. Overall, this distribution indicates that SEE emphasizes goal decomposition and long-horizon, multi-screen procedures rather than shallow single-intent interactions, aligning with the complexity of real-world mobile applications workflows. Such a subgoal structure is relatively uncommon in prior GUI datasets, and later experimental results further show that this increased compositional complexity makes fine-grained grounding the primary bottleneck in long-horizon GUI interaction.

\noindent \textbf{GUI Elements.} Moreover, one central motivation of SEE is to target modern, element-rich mobile apps.
As shown in Table~\ref{tab:ui_complexity_simple}, we quantify page complexity by the average number of interactive UI elements per screen, using a unified UI parser for fair comparison. SEE reaches 24.63 elements per screen, substantially higher than the compared datasets. This result indicates that SEE trajectories are collected from visually crowded screens where grounding is intrinsically more challenging.

\noindent \textbf{Coverage.}
Figure~\ref{fig:coverage_and_types} further illustrates the diversity of SEE and summarizes the application categories covered by the dataset, showing that SEE is built from diverse, fast-evolving real-world mobile apps. The right part presents the semantic composition of high-level instructions. Because a single instruction may contain multiple subgoals, we report multi-label coverage: each percentage denotes the fraction of episodes that contain at least one subgoal of a given type. The distribution suggests that SEE captures realistic mobile workflows involving navigation, settings adjustment, search, booking, shopping, login, payment, and related multi-step behaviors.

Overall, these statistics show that SEE complements existing GUI datasets by jointly covering long-horizon trajectories, high-complexity screens, and diverse real-world applications, while retaining multi-level supervision.

\begin{table}[]
\centering
\small
\setlength{\tabcolsep}{6pt}
\begin{tabular}{lccc}
\toprule
\textbf{Setting} & \textbf{SR} $\uparrow$ & \textbf{Type} $\uparrow$ & \textbf{Grounding} $\uparrow$ \\
\midrule
Base & 62.61\% & 93.04\% & 60.79\% \\
Base + context & 64.39\% & 94.83\% & 63.29\% \\
\midrule
FT on SEE-Train & 77.29\% & 97.87\% & 69.91\% \\
FT on SEE-Train, + context & 77.96\%  & 98.14\% & 71.85\% \\
\midrule
\textbf{$\Delta$ Context (no FT)} & 1.78\% & 1.79\% & 2.50\% \\
\textbf{$\Delta$ Context (FT)} & 0.67\% & 0.27\% & 1.94\% \\
\bottomrule
\end{tabular}
\caption{Effect of fine-tuning and retrieved context. $ \Delta$ rows report absolute improvements.}
\vskip -0.25in
\label{tab:ft_context}
\vspace{-4mm}
\end{table}

\section{Experiment}
\label{sec:exp}



We evaluate SEE from multiple complementary perspectives. First, we use SEE as a benchmark to examine whether it exposes realistic failure modes of current GUI agents on long-horizon, element-rich mobile tasks. Second, we study whether the synthesized trajectories improve generalization to unseen screens and tasks through fine-tuning and few-shot context. Third, we evaluate cross-benchmark generalization on a public mobile-control benchmark to test whether the interaction knowledge learned from SEE can be transferred beyond the training domain. Fourth, we assess the quality of the constructed transition graph through graph-quality analysis and node classification evaluation, since the graph is the structural foundation of the entire synthesis pipeline. Finally, we analyze exploration efficiency to examine whether SEE can discover deep and meaningful app functionalities with a limited interaction budget.

\begin{table*}[ht]
\centering
\setlength{\tabcolsep}{6pt}
\renewcommand{\arraystretch}{1.02}
\begin{tabular}{@{}l c ccc ccc@{}}
\toprule
\multirow{2}{*}{Model} & \multirow{2}{*}{Model Size} 
& \multicolumn{3}{c}{AndroidControl-Low} 
& \multicolumn{3}{c}{AndroidControl-High} \\
\cmidrule(lr){3-5} \cmidrule(lr){6-8}
& & Type & Grounding & SR & Type & Grounding & SR \\
\midrule
UI-Genie (NeurIPS 2025) & 3B & 97.8 & 94.7 & 93.8 & 82.5 & 82.5 & 72.9 \\
UI-Genie (Train on SEE-Train)  & 3B & 98.0 & 94.5 & 93.8 & 83.2 & 82.9 & 73.7 \\

\midrule

UI-Genie & 7B & 98.1 & 94.9 & 94.3 & 83.5 & 82.9 & 74.2 \\
UI-Genie (Train on SEE-Train)  & 7B & 98.3 & 94.2 & 94.1 & 85.2 & 82.7 & 75.7 \\

\midrule
     

Qwen2.5-VL                  & 7B & -- & -- & 91.4 & -- & -- & 60.1 \\
Qwen2.5-VL (Train on SEE-Train)         & 7B & 97.8 & 90.8 & 91.9 & 71.5 & 69.8 & 65.9 \\
\midrule
Qwen3-VL                  & 4B & 87.5 & 85.5 & 77.7 & 73.2 & 73.5 & 60.1 \\
Qwen3-VL (Train on SEE-Train)         & 4B & 89.1 & 84.2 & 79.2 & 79.7 & 75.1 & 67.8 \\
\bottomrule
\end{tabular}
\caption{Cross-benchmark transfer results on AndroidControl under low-level and high-level settings.}
\label{tab:cross_dataset}
\vspace{-4mm}
\end{table*}

\begin{table*}[ht]
\centering
\begin{tabular}{l l c c c}
\toprule
\multirow{2}{*}{Scale} & \multirow{2}{*}{Setting}
& \multicolumn{3}{c}{Graph Quality} \\
\cmidrule(lr){3-5}
& & Incorrect Edge Rate $\downarrow$ & Correct Edge Rate $\uparrow$ & Uncertain Rate $\downarrow$ \\
\midrule

\multirow{2}{*}{Small}
& Without Reflection & 27.1 & 72.1 & 0.8\\
& With Reflection    & 9.1 & 90.3 & 0.6 \\
\midrule

\multirow{2}{*}{Medium}
& Without Reflection & 29.6 & 68.7 & 1.7 \\
& With Reflection    & 11.1 & 86.6 & 2.3 \\

\bottomrule
\end{tabular}
\caption{Effect of reflection on graph quality under different evaluation scales.}
\label{tab:graph_quality_multi_scale}
\vspace{-7mm}
\end{table*}

\subsection{Experimental Setup}
\label{sec:exp_setup}

Following prior work, we report the following three evaluation metrics: Type Accuracy, Grounding Accuracy, and Success Rate (SR). Type Accuracy measures whether the predicted action type is correct. Grounding Accuracy evaluates whether the agent selects the correct UI element. Success Rate measures whether the predicted action is successfully executed.

\subsection{Agents Benchmark on SEE} 
\label{fine-tune experiement}

We first benchmark representative GUI agents on SEE to assess the realism and difficulty of our synthesized trajectories. From Table~\ref{tab:main_agents}, we observe a consistent gap between Type Accuracy and Grounding Accuracy. Many agents can infer the right operation type, yet still fail to execute the correct step because they cannot locate the correct target on element-dense screens.

This pattern is expected in modern mobile apps, where visually repetitive icons, dense lists, and near-duplicate UI elements create a large and ambiguous action space. Consequently, SR is dominated not only by planning or action selection, but by precise element-level grounding under complicated and unfamiliar layouts. 

The action breakdown accuracies further reveal failure modes. Action Swipe and Back are typically easier for strong agents because they are structurally constrained. In contrast, action Tap is the most challenging operation for most baselines, since it requires selecting the correct element among many visually similar candidates.
We also observe that Type(text) accuracy can be relatively high when the dataset provides clear instructions about what should be typed.

\subsection{Agents Few-shot Experiment}

We next assess whether SEE provides generalization benefits beyond zero-shot execution. As shown in Table~\ref{tab:ft_context}, adding retrieved context improves SR, Type, and Grounding even without parameter updates, indicating that SEE contains reusable procedural patterns that transfer to unseen episodes. The improvement is larger for Grounding than for Type, suggesting that demonstrations help disambiguate target selection on element-rich screens by exposing typical affordances and interaction patterns in similar UI contexts. Combining fine-tuning with ``+context'' yields further gains, particularly on Grounding, though the marginal benefit is smaller than in the non-fine-tuned setting, implying that part of the demonstration benefit has already been internalized during fine-tuning. Overall, these results indicate that SEE captures a key bottleneck of mobile GUI automation—fine-grained grounding on dense screens—while providing executable, step-aligned supervision that improves robustness beyond seen trajectories and layouts.

\subsection{Cross-benchmark Generalization}


We further evaluate whether SEE provides transferable supervision beyond the training domain by comparing each base model’s original performance on AndroidControl with its performance after training on SEE. As shown in Table~\ref{tab:cross_dataset}, training on SEE consistently improves Type Accuracy for almost all models under both settings, suggesting that SEE provides reusable supervision for learning action semantics and procedural structure across benchmarks. Grounding Accuracy, however, can occasionally decrease, likely due to visual domain differences between SEE and AndroidControl that make element-level localization harder even when the action type is correct. These gains are more consistent on the more challenging AndroidControl-High setting, where only coarse-grained task descriptions are available, indicating that SEE is particularly useful for long-horizon reasoning under weaker task specification. One possible reason is that SEE provides aligned median-level subgoal supervision, which may help models better decompose complex tasks and maintain progress over longer procedures.

\begin{table}[]

\centering
\small
\setlength{\tabcolsep}{7pt}
\begin{tabular}{lccc}
\toprule
\textbf{Method} & \textbf{Small} $\uparrow$ & \textbf{Medium} $\uparrow$ & \textbf{Large} $\uparrow$  \\
\midrule
OCR Matching     & 90.5\% & 83.3\% & 58.1\% \\
Pixel Detection & 92.3\% & 80.4\% & 65.2\%  \\
Icon Feature Detection & 95.7\% & 90.6\% & 75.4\%  \\
SEE                  & 99.3\% & 98.6\% & 96.9\%  \\
\bottomrule
\end{tabular}
\caption{Classification accuracy of screen-node matching methods under different graph scales.}

\label{tab:screen_classification_scales}
\vskip -0.3in
\end{table}

\begin{table}[]
\centering
\small
\setlength{\tabcolsep}{6pt}
\begin{tabular}{c|ccc}
\toprule
\textbf{Coverage target} &
\textbf{Random} $\downarrow$ &
\textbf{LLM} $\downarrow$ &
\textbf{Ours} $\downarrow$ \\
\midrule
20\%  & 105 & 55 & 59 \\
60\%  & -- & 375 & 208 \\
80\%  & -- & -- & 397 \\
\bottomrule
\end{tabular}
\vspace{1mm}
\caption{Average steps to cover core functionalities on DingTalk.
The missing entries (``--'') indicate that across repeated trials it failed to reliably reach the corresponding coverage target within limited budget.
}
\vskip -0.3in
\label{tab:dingtalk_doc_coverage}
\end{table}

\subsection{Graph Quality Study}
\label{sec:graph quality}

To evaluate the effectiveness of the reflection mechanism on graph quality, we design an edge-level evaluation protocol based on an independent LLM auditor~\cite{qwen}. For a fair comparison, both the reflected and non-reflected views are derived from the same exploration run. During exploration, all discovered transitions are added into the graph, while the reflection module only attaches review labels and does not modify edges online. Therefore, both the reflected and non-reflected views are derived from the same graph and the same set of explored transitions, eliminating confounding factors caused by different exploration trajectories or coverage.

Based on this setup, we sample transition edges from the constructed graph and ask a separate multimodal LLM auditor to judge whether each edge is semantically and functionally valid. We report results under two testing scales: a small-scale setting with about 500 edges and a medium-scale setting with about 1,000 edges.

As shown in Table~\ref{tab:graph_quality_multi_scale}, reflection consistently improves graph quality at both scales. The benefit remains stable as the graph grows. We attribute this robustness to the design of the reflection module. Reflection is performed after each executed action and evaluates whether the observed outcome is consistent with the local action context. Because this decision is made locally, its effectiveness is only weakly affected by the overall graph scale.


\paragraph{Node Classification Evaluation} To further assess the robustness of our screen-node classification module, we conduct a dedicated quantitative evaluation. We compare SEE with three global screen-matching baselines: an OCR-based~\cite{cui2025paddleocr30technicalreport} method, a pixel-difference method, and an icon-feature-difference method. The comparison is conducted under three graph scales, small, medium, and large, corresponding to 98, 238, and 463 screen images, respectively. The ground-truth labels used in this evaluation were annotated by human judges.

All three baselines perform matching over the full interface. The OCR baseline uses global text similarity and classifies two observations as different when the difference exceeds 0.35. The pixel-difference baseline compares whole-screen visual features and treats two observations as the same state when their similarity is above 0.90. The icon-feature baseline encodes all detected icons and compares the aggregated icon representations, using a similarity threshold of 0.92. For fairness, SEE and the icon-feature baseline share the same icon encoder, MobileNetV3\cite{howard2019searching}.

Table~\ref{tab:screen_classification_scales} shows that SEE consistently outperforms all baselines across all graph scales. The results suggest that SEE is substantially more robust to scale growth, reducing erroneous state splitting and helping preserve a compact and reliable transition graph in large-scale exploration. 


\subsection{Exploration Efficiency Study}
A key motivation of SEE is to efficiently discover deep and prerequisite-dependent UI states during exploration, so that the subsequent synthesis stage can compose realistic long-horizon workflows. To evaluate this property, we define functional coverage based on an app’s official documentation. Because DingTalk~\cite{dingtalk} provides a relatively complete description of its functionalities, we use it as a testbed for this analysis. Specifically, we collect a set of core functionalities from the official documentation and measure how many interaction steps are required to reach different coverage targets. 

Table~\ref{tab:dingtalk_doc_coverage} compares random exploration, an LLM-only \cite{qwen} exploration, and our proposed SEE. Result shows that SEE reaches 20\% coverage with about 60 steps and scales to 80\% with about 400 steps on average, indicating that our structured exploration can penetrate meaningful functional regions efficiently. 
In contrast, random exploration requires substantially more steps even for shallow coverage and often fails to reach higher coverage targets, and the LLM-only baseline becomes unstable as the target coverage increases.
Overall, these results show that SEE is effective at reaching deep functional regions of an app.


\section{Conclusion}
We presented SEE, a structure-aware data synthesis framework and dataset for training and evaluating GUI agents in realistic, mobile applications. The resulting dataset emphasizes long trajectories, dense screens, and aligned multi-level supervision, providing a challenging yet practical testbed for modern mobile GUI agents. Experiments show that SEE exposes realistic failure modes of current agents, improves performance under finetuning and retrieval-based prompting. In future work, we plan to extend SEE to more complex instruction settings and synthesize longer-horizon trajectories. We also plan to develop a difficulty-aware scoring and stratification scheme for trajectories, so as to support more future research. 

\section{Acknowledge}
This work is partially supported by the National Natural Science Foundation of China under Grants 62441232, 62476068, 62306092, 62502115, and projects ZR2025ZD01, ZR2024QF066, ZR2025QC1516 supported by Shandong Provincial Natural Science Foundation.

\bibliographystyle{ACM-Reference-Format}
\bibliography{sample-base}

\clearpage
\appendix
\input{supplementary}










\end{document}

%% file: supplementary.tex
\section{Case Study}
\label{Case study}
Our graph-based synthesis over an explicitly discovered screen--element transition graph enables two types of
behaviors that are rarely covered by existing GUI datasets: \textbf{(i) deep corner trajectories} that reach
low-frequency, prerequisite-dependent screens (e.g., identity verification, service agreements), and
\textbf{(ii) interruption-handling trajectories} that preserve task context under execution-time UI perturbations
(e.g., phone calls, pop-up overlays). Prior datasets often discard such episodes as ``noise'', yet they are
ubiquitous in real mobile interaction and constitute a major failure source for deployed GUI agents.
Below we present some examples from SEE, highlighting long-horizon task composition, corner-case coverage, and context-preserving recovery behaviors.

\begin{figure}
    \centering
    \includegraphics[width=0.9\linewidth]{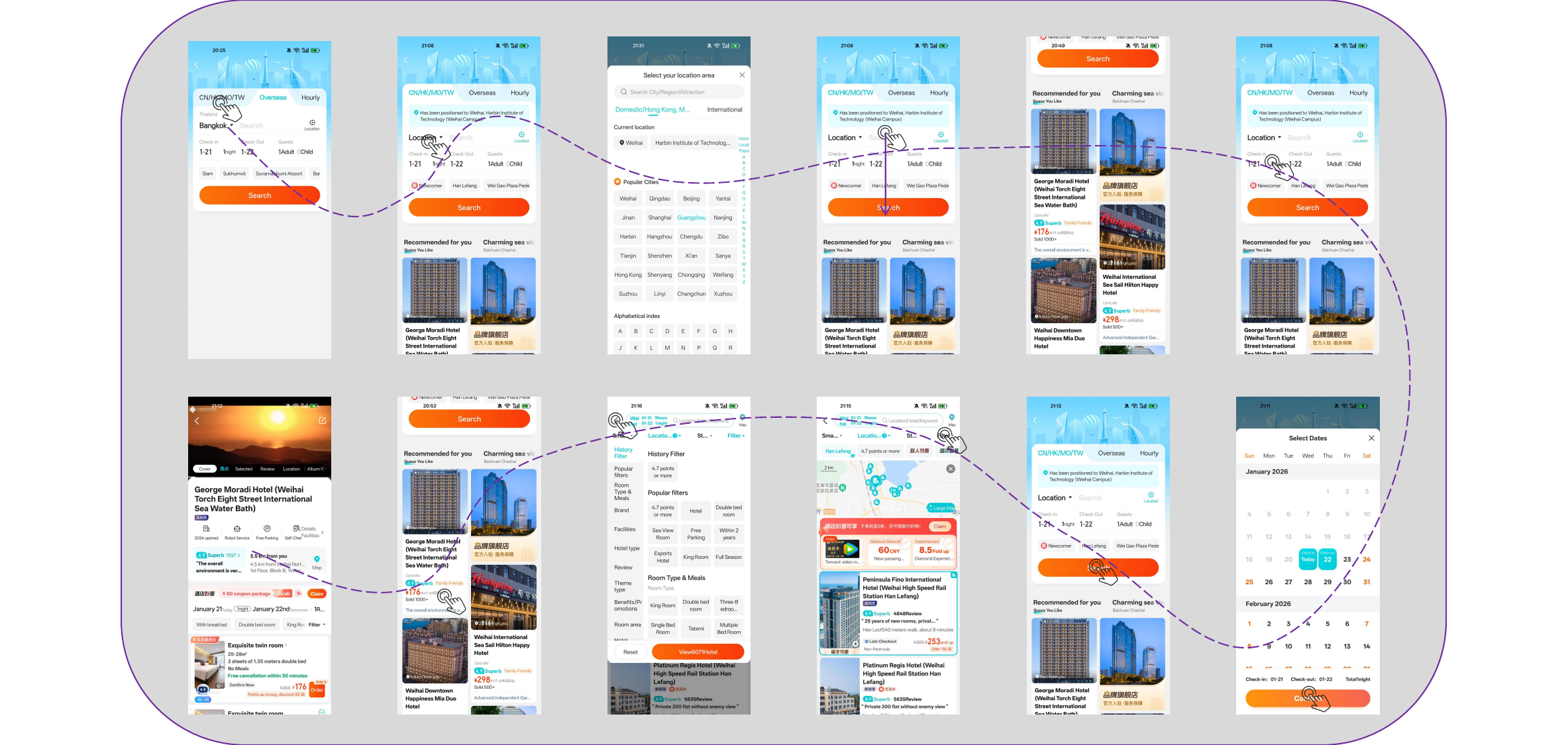}
    \caption{\textbf{A long-horizon, executable trajectory with step-aligned supervision.} 
    The episode starts from an international hotel search interface and composes a multi-step booking workflow,
    including destination selection, date picking, search refinement, and drilling down to a specific hotel detail page.
    The provided supervision is \emph{multi-level} and \emph{step-aligned}: a high-level query specifies the overall goal,
    subgoals decompose the task into meaningful milestones, and each step includes a low-level instruction describing
    the intended operation and expected screen transition.}
    \label{fig:Example1 from dataset}
\end{figure}

Figure~\ref{fig:Example1 from dataset} shows an episode in SEE.
Besides the executable action trace, each episode provides a high-level query and an explicit set of
subgoals for long-horizon decomposition. The \textbf{high-level query} is: Plan a hotel booking journey starting from an international search screen, progressing through location selection, date picking, search refinement, and viewing detailed hotel information. And \textbf{subgoals} are: (1) Navigate to location selection screen to choose a destination, (2) Select Weihai as the travel destination, (3) Open date picker to set check-in and check-out dates, (4)Apply filters to refine hotel search by price and ratings, (5) View detailed information about a specific hotel option.

\begin{figure}[t]
    \centering
    \includegraphics[width=\linewidth]{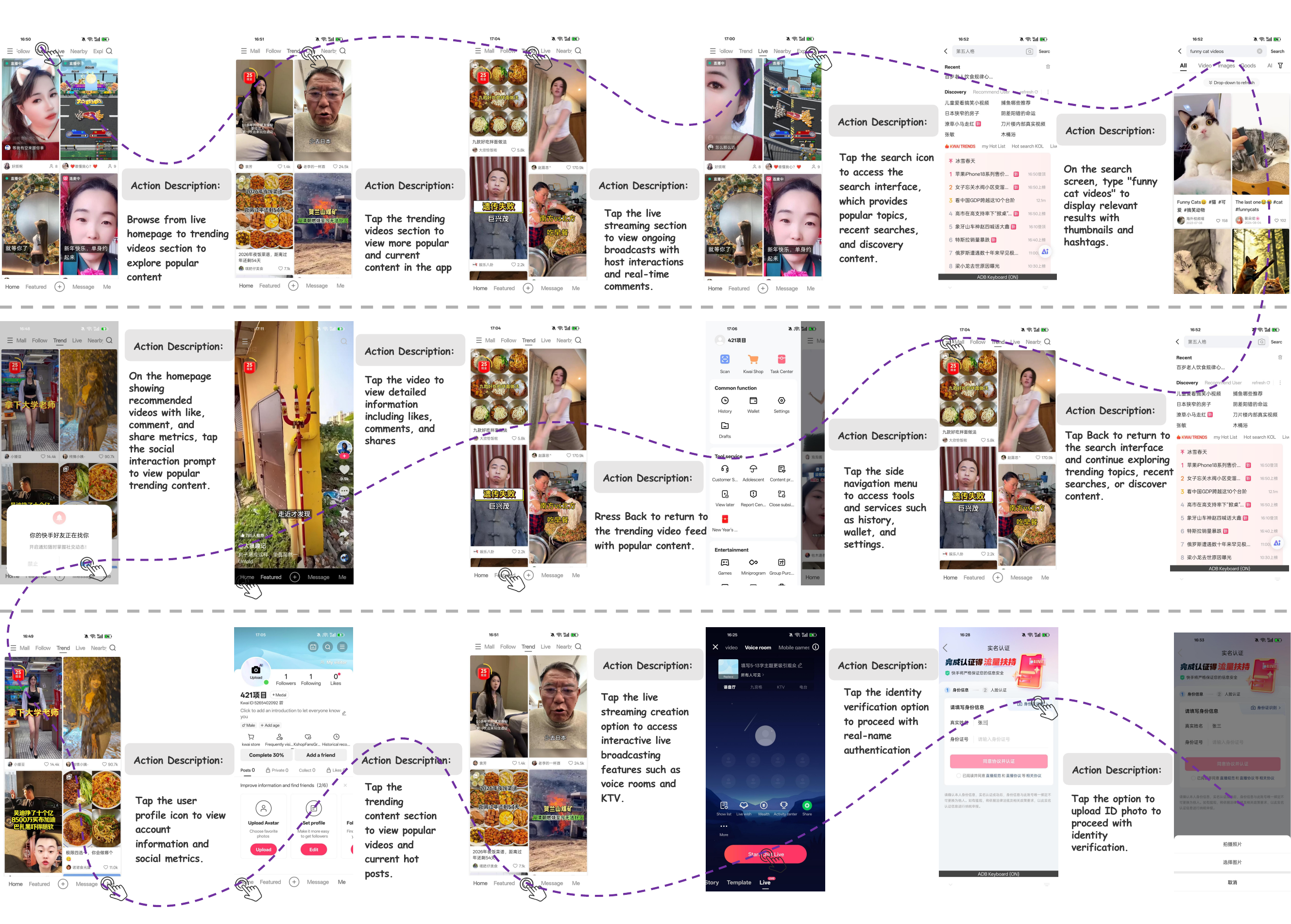}
    \caption{Example long-horizon trajectory in SEE with corner cases and interruptions.  The trajectory covers both common browsing/navigation flows (e.g., trending/live/search) and rare, prerequisite-dependent
states such as identity verification, demonstrating SEE’s ability to reach deep corner interfaces.
Moreover, the episode includes transient interruptions (e.g., pop-ups/overlays) without dropping context: the instructions remain consistent with the ongoing goal and continue to specify precise operations and targets despite UI changes.}
    \label{fig:Long and Corner Example}
\end{figure}

Figure~\ref{fig:Long and Corner Example} illustrates a more typical user trajectory: although the user goal is
coherent, the execution path may involve detours, deep screens.
Such trajectories are difficult to obtain from naive online exploration and are often missing from prior datasets, yet they are crucial for training robust GUI agents.
In this example, the \textbf{high-level instruction} is: Navigate from the live streaming home page to complete real-name authentication via a structured user journey involving discovery, search, profile access, and identity verification. And \textbf{subgoals} are: (1) Browse from live homepage to trending videos section to explore popular content. (2) Access search interface to look for specific content or services. (3) Navigate to side menu to access account and security settings. (4) Open user profile to initiate identity verification process. (5) Enter real-name authentication page to begin submission of identity information. (6) Proceed to upload ID document through photo capture or selection interface. As we can see from the trajectory, the whole trajectory accomplish the overall step by step, provides concrete low-level instruction at the same time. 

\begin{figure}[htbp]
    \centering
    \includegraphics[width=1.0\linewidth]{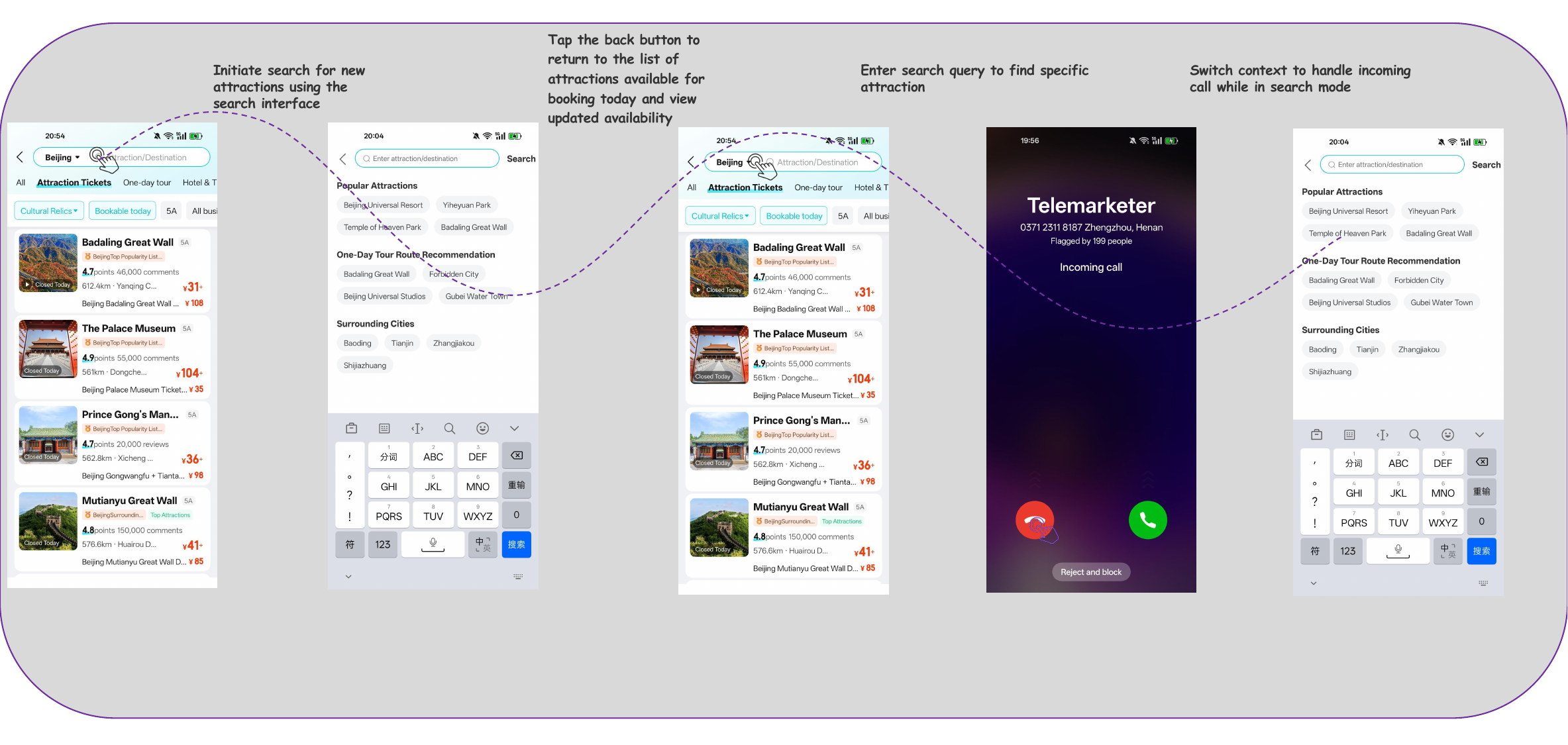}
    \caption{Interrupted by phone call. In the trajectory, our framework recognize the phone call interruption and decide to handle the phone call while keeping the trajectory context. }
    \label{fig:interruption phnoe call example}
\end{figure}

Here we show another example that handles more direct interruption \ref{fig:interruption phnoe call example}, which is a pretty common in real interaction scenarios.
In real agent deployment scenarios, agent execution can be interrupted by phone call or notice, pop up window. Existing usually lack these data, or filter these bad cases out. In SEE, we treat these trajectories and normal, and generate decent description for them.

\section{Exploration Phase study}
\label{app:exploration phase}

A practical challenge in real-world mobile apps is that many screens are \emph{content-dynamic} while remaining the same functional state in the navigation structure.
For example, in shopping applications, the cart page may change substantially when items are added/removed (e.g.,
list length, prices, and counters), but it should still be treated as the same screen node for building a faithful transition graph.
This type of dynamic update often breaks screen matching strategies that rely on global appearance differences or
content-heavy structures, leading to spurious node splits and noisy transitions.

\begin{figure}[htbp]
    \centering
    \includegraphics[width=\linewidth]{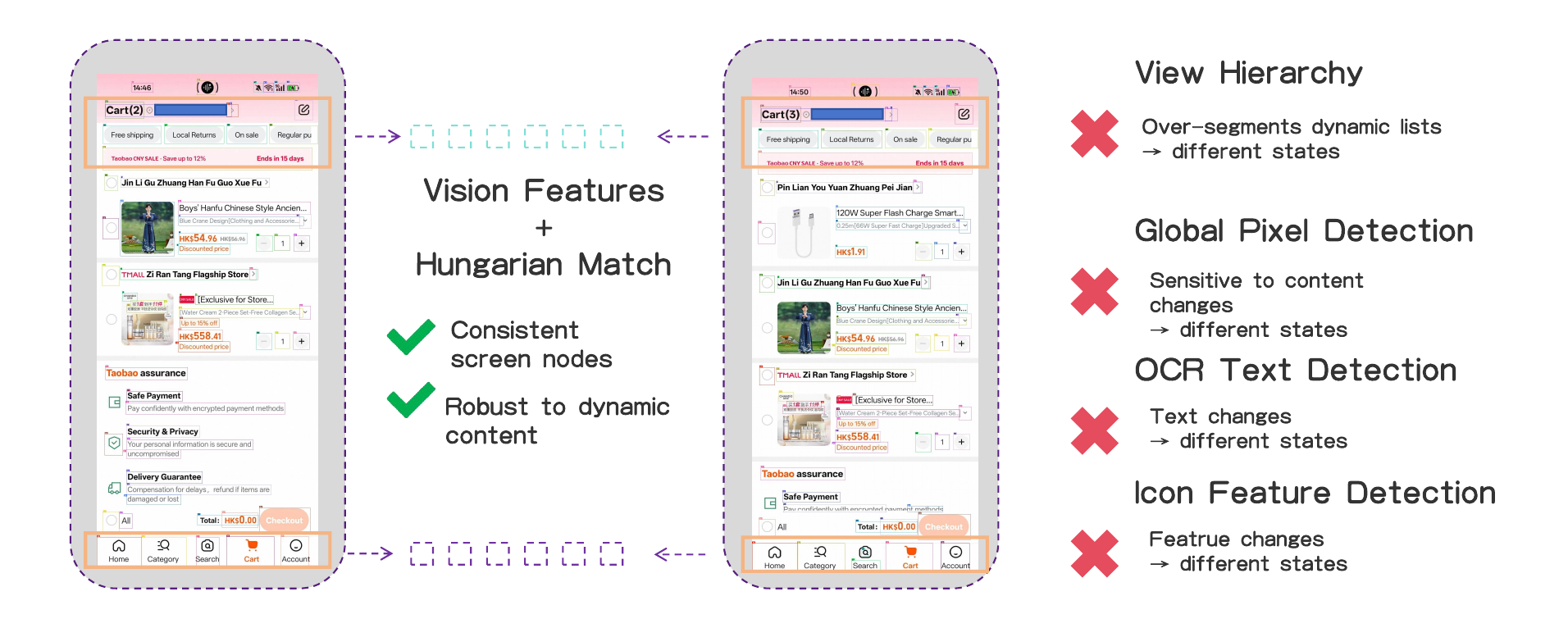}
    \caption{\textbf{Screen identification case study on a dynamic cart page.}
    The two observations correspond to the same cart screen before and after adding an item (list content changes).
    Our method uses stable-region visual prototypes (e.g., top/bottom bars) and Hungarian matching to classify them into the same screen node, preserving the true transition structure.
    Baselines relying on view-hierarchy, global pixel change detection, frame-level feature detection or OCR/text matching are brittle to content updates and may incorrectly split a single functional screen into multiple nodes.}
    \label{fig:screen_id_case}
    \vspace{-2mm}
\end{figure}

Figure~\ref{fig:screen_id_case} shows a representative example from the cart page. After adding one item, the cart content changes, yet our screen matches method via Hungarian assignment and correctly merges the two
observations into the same screen node. In contrast, several commonly used heuristics would incorrectly classify them as different states. This robustness is important for preventing spurious nodes/edges and for enabling reliable graph-based trajectory synthesis on modern, element-rich apps.

\begin{figure}[htbp]
    \centering
    \includegraphics[width=1.0\linewidth]{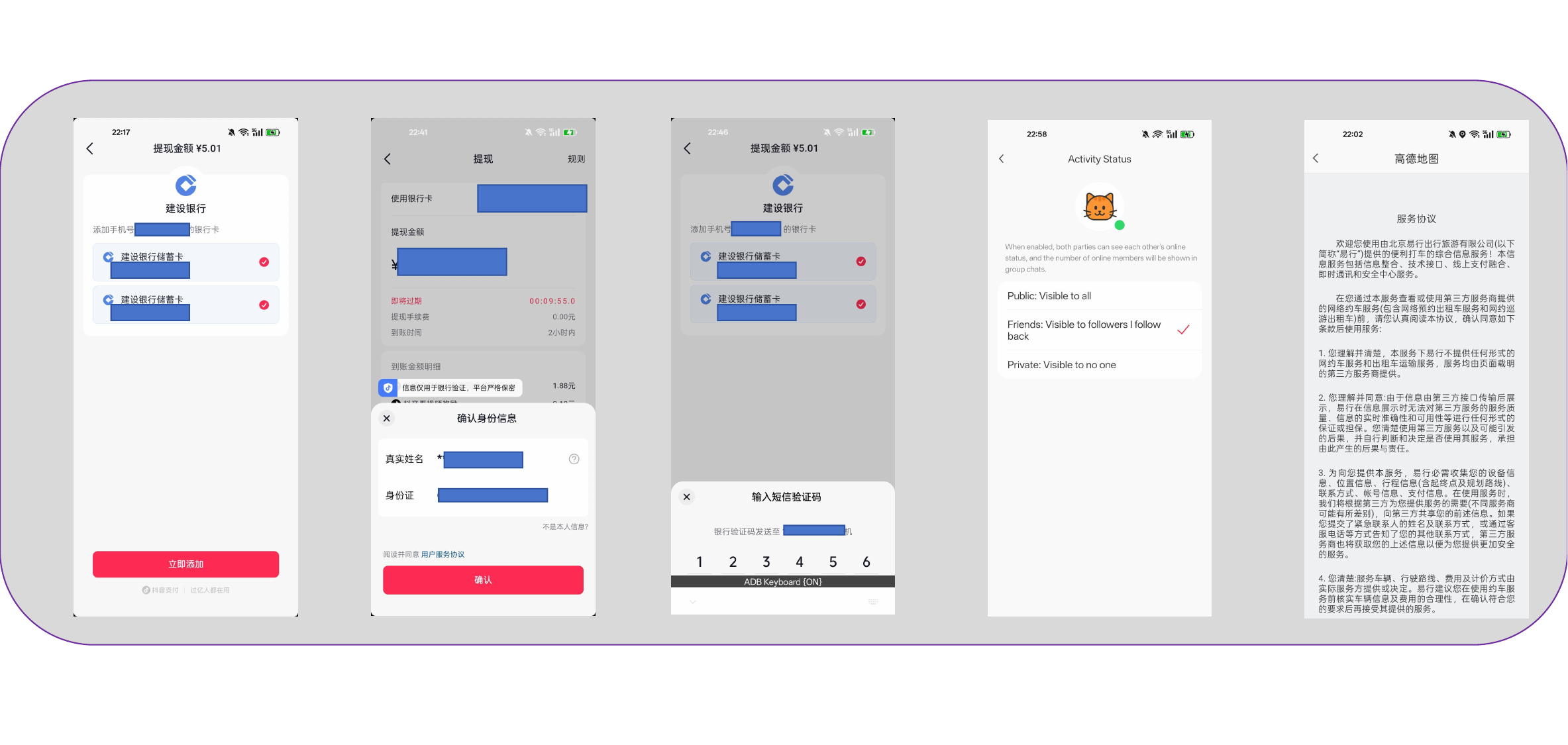}
    \caption{Examples of deep, low-frequency corner-case page. \textbf{Left (three panels)}: a multi-step withdrawal workflow in Douyin, reaching security-critical screens such as bank-card selection and identity verification (real-name/ID fields) and the SMS code input overlay—states that are rarely covered by typical demonstrations due to strict prerequisites and transient dialogs. \textbf{Middle-right}: a privacy-related “Activity Status” setting in Red Notes, illustrating navigation into fine-grained account visibility controls. \textbf{Rightmost}: Amap’s service agreement/terms-of-use consent page, representing long-form legal pop-ups that commonly interrupt workflows. These cases highlight that our synthesized trajectories can traverse deep UI states and handle prerequisite-dependent dialogs and pop-up interfaces, substantially expanding coverage beyond common “happy-path” flows.}
    \label{fig:deep corner pages}
\end{figure}

Moreover, Figure \ref{fig:deep corner pages} displays some very deep and corner pages within apps. Our structure-aware exploration strategy can lead to very deep screens, offers more potential for future mobile agent development. These function page are even rarely visited by human demonstration, but they are curcial for future general GUI Agent development. 

\section{Failure Cases Discussion On SEE Dataset}
\label{sec:appendix_failures}

We discuss representative failure cases to clarify why several evaluated GUI agents exhibit modest success rates on our dataset. Overall, failures are dominated by two dataset characteristics that reflect real-world mobile apps: \textbf{element-dense screens} and \textbf{interruption situation in long task chains}. These factors interact to amplify grounding errors and cause compounding deviations over time.

\begin{figure}
    \centering
    \includegraphics[width=0.9\linewidth]{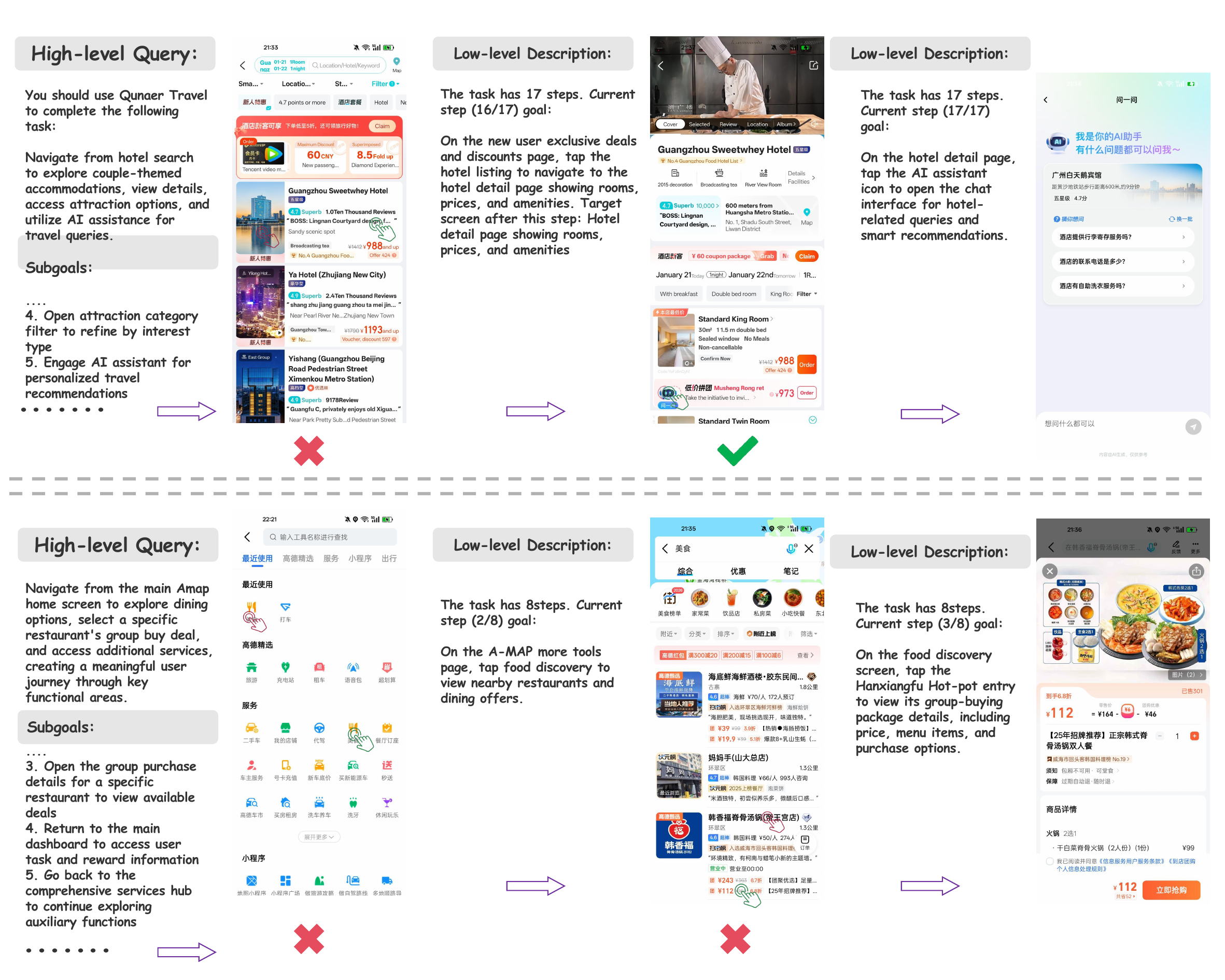}
    \caption{\textbf{Representative grounding failures caused by dense and repetitive UI elements.}
Green markers denote the ground-truth (GT) target position, while red markers denote the model’s predicted tap position. \textbf{Top:} the model taps the hotel tab icon whereas the GT taps the corresponding text label; although both are semantically consistent with the instruction, they are treated as different targets under element-level evaluation.
\textbf{Bottom:} the model taps a \emph{``Food''} entry that is visually and semantically correct, but the screen contains multiple ``Food'' icons; the prediction selects a different instance than the GT and is therefore counted as incorrect. In the second step, the low-level instruction is tap the group-buy package, but model can't recognize the group-buy meaning, therefore tap the restaurant icon instead.
These cases illustrate that, on element-rich screens, small localization ambiguities, duplicated affordances and complex UI page can dominate error patterns even when the intended action is correct.}
    \label{fig:failure cases}
\end{figure}

\paragraph{Element-rich UI increases grounding ambiguity.}
Many screens in our dataset contain a large number of interactive elements with high visual and semantic similarity. In some cases, there are some identical elements in one page, which might lead to multiple real answers in on time step \textit{t}. The figure \ref{fig:failure cases} displays how complex interactive elements affects grounding accuracy

Even when the agent correctly infers the intended action type (e.g., \texttt{Tap} vs.\ \texttt{Swipe}) from our
accurate low-level descriptions, it must still localize the correct target among many plausible candidates.
In practice, small perceptual differences, partial occlusions, and layout variations often cause the agent to click a nearby but incorrect element, resulting in an unintended state transition. This explains the common gap observed between
high Action Type Accuracy and comparatively lower UI Element Interaction Accuracy (Grounding) in the main paper.

\paragraph{Why accurate low-level descriptions are not sufficient.}
Our dataset provides precise step-level descriptions that specify \emph{what} action should be taken and often
\emph{which} element is intended at a semantic level. However, mapping this semantic intent to the correct pixel-level
target remains challenging in element-dense screens, especially for icon-only controls, transient pop-ups, and repeated
UI patterns. Thus, the main bottleneck is not action planning, but reliable grounding and state maintenance under
realistic UI complexity.

\paragraph{Implication.}
These failures suggest that our dataset is well-specified (agents can often predict the correct operator) yet
appropriately challenging (grounding and long-horizon robustness remain limiting factors). It therefore serves as a stress test for GUI agents in modern, fast-evolving applications, and motivates future work on stronger grounding, uncertainty-aware action selection, and recovery policies.

\paragraph{Ablation Study on Exploration Efficiency}

To complement the exploration-efficiency analysis, we further conduct an ablation study on the exploration policy. Specifically, we evaluate the contribution of three designed components: semantic prior, layout prior, and history penalty. For each variant, we measure the average number of interaction steps required to reach different coverage targets, following the same protocol as in the main text.

\begin{table*}[h]
\centering
\small
\setlength{\tabcolsep}{4.5pt}
\begin{tabular}{lccccccc}
\toprule
\textbf{Variant} & \textbf{Semantic} & \textbf{Layout Prior} & \textbf{History Penalty} & \textbf{20\%} $\downarrow$ & \textbf{60\%} $\downarrow$ & \textbf{80\%} $\downarrow$ \\
\midrule
w/o Semantic           &  & \checkmark & \checkmark  & 78 & 276 & 562 \\
w/o Layout Prior     & \checkmark &  & \checkmark  & 83 & 354 & 594 \\
w/o History Penalty    & \checkmark & \checkmark &   & 99 & 398 & -- \\
\midrule
Full SEE Exploration   & \checkmark & \checkmark & \checkmark  & 59 & 208 & 397 \\
\bottomrule
\end{tabular}
\caption{Ablation study on exploration efficiency. }
\label{tab:ablation_exploration_efficiency}
\end{table*}

Table~\ref{tab:ablation_exploration_efficiency} shows that removing any of these components reduces exploration efficiency, indicating that each of them contributes positively to the exploration process. Without the semantic prior, the agent is less capable of selecting functionally meaningful elements. Without the layout prior, it becomes harder to exploit common interface structure for efficient navigation. Without the history penalty, the agent is more likely to revisit previously explored or unproductive interactions. Overall, the full SEE exploration strategy performs best across different coverage targets, showing that these components are complementary and jointly important for efficient exploration.

\section{Prompts Used in Exploration}
\label{app:prompts}

\paragraph{Action selection prompt (Exploration).}
We use the following prompt to let the LLM choose exactly one interaction on the current screen given the parsed UI element list. The prompt is instantiated with the app name, screen size, ranked element candidates. Notice that in our exploration stage, the \texttt{Back} action is executed in a rule-based manner to ensure the exploration can progress autonomously: when the agent appears to be trapped in an overly deep screen that yields no new transitions (e.g., staying on the same screen for more than $n$ consecutive steps), we trigger \texttt{Back} to return to a higher-level state and continue discovering new branches.
In contrast, during synthesis, \texttt{Back} is strongly context-dependent and is generated as part of the planned path on the transition graph, serving as a necessary step to reach reachable target nodes rather than a generic recovery heuristic.

\begin{promptbox}[Element-selection prompt]

\small\ttfamily
You are a helpful UI exploration agent. Your task is to explore the function of the \{app\} as much as possible.
You will be given some top candidate icons. Your task is to pick one element that you think is most likely to help you achieve this goal.
Consider the following factors:
\begin{itemize}
    \item The icon's relevance to the app's main functionality and navigation.
    \item The icon's position on the screen, with preference for those located in typical navigation areas (e.g., top or bottom bars).
    \item The icon's novelty, favoring icons that have not been clicked frequently in the past to encourage diverse exploration.
    \item The icon's historical effectiveness, avoiding icons that have previously led to no state changes.
\end{itemize}

From the following list of top candidate icons, select the one that best aligns with these criteria.
Respond with the index of the chosen icon in the provided list.

\vspace{2pt}
\textbf{Top Candidate Icons:}
\begin{itemize}
    \item 0: \{icon\_text\_0\}
    \item 1: \{icon\_text\_1\}
    \item $\cdots$
    \item K-1: \{icon\_text\_{K-1}\}
\end{itemize}

\textbf{Tapped elements history:} \{history\_list\}

Please provide only the index number of the selected icon.
Your answer should be in JSON format:
\begin{center}
\texttt{\{"selected\_index": index\_number\}}
\end{center}

You are recommended to choose the element that can complete a meaningful task based on the history.
For example, if the history shows you have just opened a settings page, consider choosing elements that help you explore settings options.
Avoid choosing elements that lead to no state change.
\label{prompt: selection}
\end{promptbox}

Other prompts used in exploration are as followed. Exploration prompt is used to select element from the ordered candidate elements.Transition record reflection is used to generate reflection during exploration.

\begin{promptbox}[Exploration prompt (Exploration)]
\setlength{\parskip}{1em}   
\setlength{\parindent}{0pt}   
\#\#\# Background \#\#\#\\
You are exploring the current page of the mobile app '\{app\_name\}'. Your goal is to build a screen-transition graph for later data synthesis. At each step, you must perform exactly ONE interaction on the current screen to discover new screens and transitions.\\

\#\#\# Screenshot information \#\#\#\\
We extracted perceived elements from the screenshot via a UI parser. The screen size is width:\{width\} x height:\{height\} pixels.\\
Each line is [x, y]; content. Coordinates are pixel positions; content is either text or an icon description. Icon descriptions might be noisy and should be cross-checked against the screenshot in your mind.\\

\#\#\# Keyboard status \#\#\#\\
We extract the keyboard status of the current screenshot and it is whether the keyboard of the current screenshot is activated.\\
The keyboard status is as follows:\\
- The keyboard \{IS / is NOT\} activated ...\\

\#\#\# Exploration requirements \#\#\#\\
- The exploration policy has already selected a target element for this step.\\
- The index of the selected element is \{idx\}.\\
- The description of the selected element is: '\{element\_desc\}'.\\
- By default, you should try to interact with this element using a Tap(x, y) action that clicks near its center.\\
- Only if you believe tapping this element is clearly unhelpful for exploration, you may choose other action instead.\\
- If you choose a Swipe action, use it to reveal NEW content ... recommended swipe length is about 600 pixels.\\
- If the keyboard is activated, you are recommended to choose a Type(text) action to type a SHORT, meaningful query ...\\

\#\#\# Response requirements \#\#\#\\
You must now choose exactly ONE of the following actions and apply it to the current page:\\
- Tap (x, y): Tap the position (x, y) on the current page.\\
- Swipe (x1, y1), (x2, y2): Swipe from position (x1, y1) to position (x2, y2).\\
- Type (text): Type the given "text" into the currently focused input field ...\\

\#\#\# Output format \#\#\#\\
Your output consists of the following three parts:\\
\#\#\# Thought \#\#\#\\
Reason about which single action ... and what screen this action is expected to lead to.\\
\#\#\# Action \#\#\#\\
You can only choose one from the actions above. Make sure that the coordinates or text in the "()".\\
\#\#\# Operation \#\#\#\\
Generate a brief natural language description of the operation in the Action, consistent with your Thought.\\

\label{prompt:task_completion}
\end{promptbox}

\begin{promptbox}[Transition Record Prompt (After-Action Reflection)]

\small
\textbf{System / Background.}\\
These are two screenshots of the app ``\{APP\}'' \emph{before} and \emph{after} a single interaction.
The screen size is \{WIDTH\}$\times$\{HEIGHT\} pixels.

\vspace{0.5em}
\textbf{Before action (tap/swipe/type).}\\
We provide the parsed UI element list extracted from the \emph{before-action} screenshot.  
Each line is formatted as: \texttt{[x, y]; content}.  
\textit{(The list below is populated at runtime.)}
\begin{itemize}
    \item \texttt{\{(x\_1, y\_1); text\_1\}}
    \item \texttt{\{(x\_2, y\_2); text\_2\}}
    \item \dots
\end{itemize}

\vspace{0.5em}
\textbf{After action (tap/swipe/type).}\\
We provide the parsed UI element list extracted from the \emph{after-action} screenshot.  
Each line is formatted as: \texttt{[x, y]; content}.  
\textit{(The list below is populated at runtime.)}
\begin{itemize}
    \item \texttt{\{(x'\_1, y'\_1); text'\_1\}}
    \item \texttt{\{(x'\_2, y'\_2); text'\_2\}}
    \item \dots
\end{itemize}

\vspace{0.5em}
\textbf{Action / interacted target (may be noisy).}\\
The following is an LLM-based description of the interacted element/region.  
It may be imperfect or partially incorrect:
\begin{quote}
\texttt{\{CLICKABLE\_DESC\}}
\end{quote}

\vspace{0.5em}
\textbf{Task.}\\
Summarize the transition in \textbf{ONE} sentence that is suitable for building a graph.
Use the form:
\begin{quote}
\texttt{SourceScreen -> TargetScreen via <Action>}
\end{quote}
where \texttt{<Action>} is a concise phrase describing what the user did.  
Also provide a one-sentence summary of the \texttt{TargetScreen} semantics (what this screen is mainly about).

If the two screens show essentially the same content and layout (no meaningful change), treat them as the \textbf{SAME} screen
(i.e., no transition), and reflect this in your \texttt{Transition}.

\vspace{0.5em}
\textbf{Output format.}\\
\texttt{\#\#\# Transition}\\
\texttt{SourceScreen -> TargetScreen via <Action>}\\
\texttt{\#\#\# Target summary}\\
\texttt{One-sentence description of the target screen.}
\label{prompt: reflection}
\end{promptbox}